\pdfoutput=1

\documentclass[11pt]{article}

\usepackage[preprint]{coling}

\usepackage{times}
\usepackage{latexsym}

\usepackage[T1]{fontenc}

\usepackage[utf8]{inputenc}

\usepackage{microtype}

\usepackage{inconsolata}

\usepackage{graphicx}
\usepackage{booktabs}
\usepackage{multirow}
\usepackage{amsmath}
\usepackage{comment}
\usepackage{amsfonts}
\usepackage{amssymb}
\usepackage{tabularx}
\usepackage{adjustbox}
\usepackage{enumitem}
\usepackage{listings} 
\title{Mixture of Prompt Learning for Vision Language Models}

\author{
	Yu Du$^{\dagger}$, Tong Niu$^{\dagger}$, Rong Zhao$^{ *}$ \\
	Center for Brain-Inspired Computing Research\\
	Department of Precision Instrument, Tsinghua University \\
    \texttt{\{duyu20,nt20\}@mails.tsinghua.edu.cn}, \texttt{r\_zhao@tsinghua.edu.cn}\\
    \texttt{$\dagger$ Equal contribution}, \texttt{* Corresponding author}
}

\begin{document}
\maketitle
\begin{abstract}
As powerful pre-trained vision-language models (VLMs) like CLIP gain prominence, numerous studies have attempted to combine VLMs for downstream tasks. Among these, prompt learning has been validated as an effective method for adapting to new tasks, which only requiring a small number of parameters.
However, current prompt learning methods face two challenges: first, a single soft prompt struggles to capture the diverse styles and patterns within a dataset; second, fine-tuning soft prompts is prone to overfitting. To address these challenges, we propose a mixture of soft prompt learning method incorporating a routing module. This module is able to capture a dataset's varied styles and dynamically selects the most suitable prompts for each instance. Additionally, we introduce a novel gating mechanism to ensure the router selects prompts based on their similarity to hard prompt templates, which both retaining knowledge from hard prompts and improving selection accuracy. We also implement semantically grouped text-level supervision, initializing each soft prompt with the token embeddings of manually designed templates from its group and applied a contrastive loss between the resulted text feature and hard prompt encoded text feature. This supervision ensures that the text features derived from soft prompts remain close to those from their corresponding hard prompts, preserving initial knowledge and mitigating overfitting. 
Our method has been validated on 11 datasets, demonstrating evident improvements in few-shot learning, domain generalization, and base-to-new generalization scenarios compared to existing baselines. The code will be available at \url{https://anonymous.4open.science/r/mocoop-6387}
\end{abstract}
\section{Introduction}
Recently, pre-trained vision-language models like CLIP become increasingly prominent, numerous studies have explored their application in various downstream tasks such as image classification~\cite{zhou2022learning}, visual question answering (VQA)~\cite{eslami2021does}, and cross-modal generation~\cite{crowson2022vqgan}. Prompt learning has emerged as an effective method by optimizing the prompts fed into the model, significantly improving performance on new downstream tasks without requiring large-scale fine-tuning of the entire model. 

For example, take the downstream task of image classification, the prompt essentially serves as a template that can be positioned before, after, or surrounding the class name. Traditionally, manually designed text templates were used during the training of CLIP, guide the model in associating textual descriptions with visual content. These manually designed prompts are called hard prompts. Prompt learning takes this a step further by replacing these fixed text templates with learnable continuous vectors. By fine-tuning these vectors with a small number of samples, the performance on downstream tasks can be significantly improved. These vector-based prompts are called soft prompts to distinguish them from hard prompts.

We focus on two challenges of soft prompt learning in this work.
\begin{figure*}[t]
\centering
\includegraphics[width=1.\textwidth]{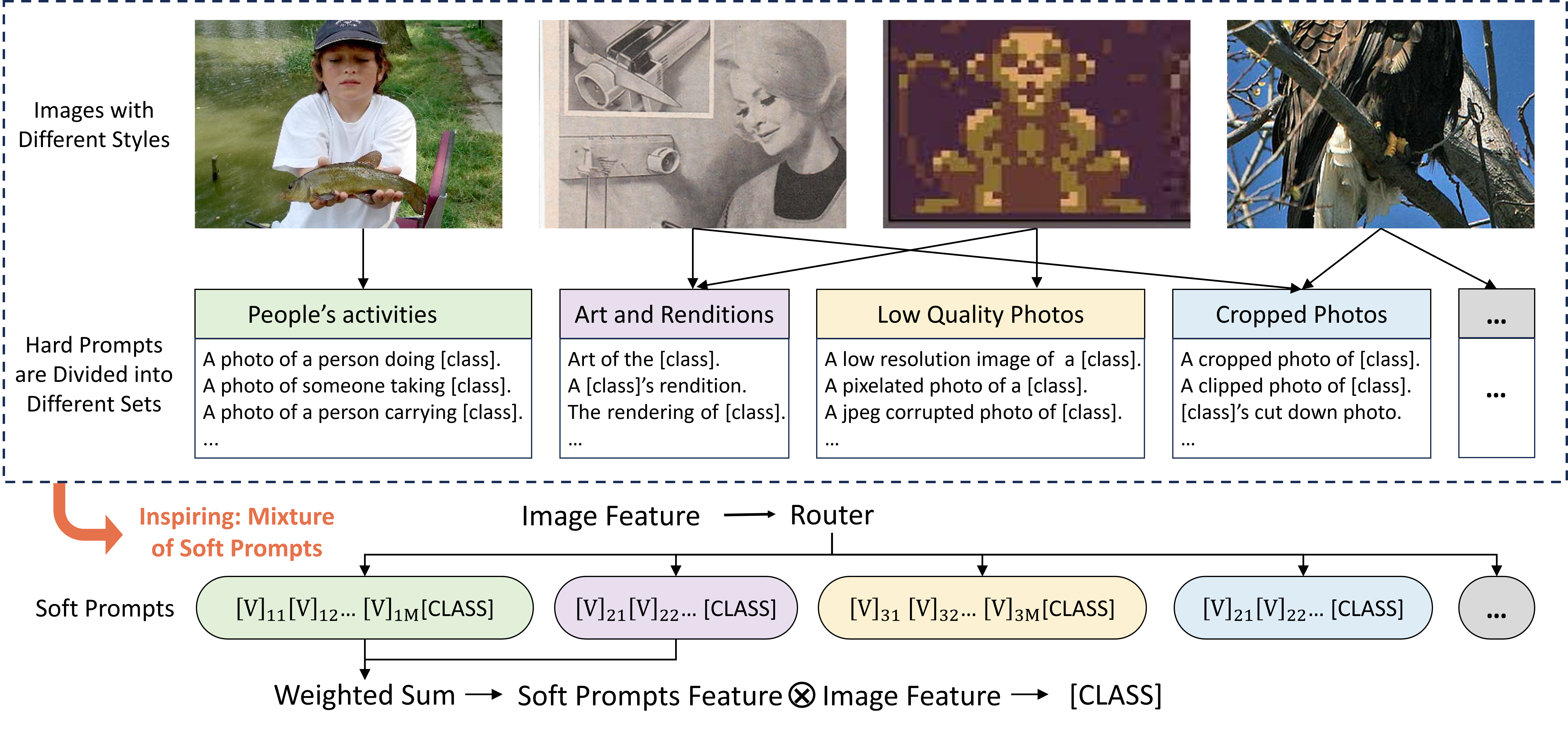} 
\caption{For a dataset, the existing hard templates can be divided into different sets based on the different styles and patterns  they describe in the images (such as different contents within the different colored blocks). Furthermore, one image can simultaneously possess multiple different styles. Traditionally, only one soft prompt is used to fit all images, but we use multiple soft prompts. Each soft prompt represents a style, and a router selects the best matches. This approach better bridges the gap between visual and text features by taking different styles into consideration. }
\label{fig:Moltivation}
\end{figure*}
\noindent{1) Dataset style variations.}
As seen in Figure.~\ref{fig:Moltivation} For one dataset, a single soft prompt may not be sufficient to capture the diverse styles present in the data. Difference instances in the same dataset may be compatible with different prompts. Therefore, it is more natural to use multiple prompts to represent these variations adequately. 
\noindent{2) Overfitting Issue.} Improper finetuning of the soft prompts may result in performance that even lags behind the zero shot capabilities of the original VLMs~\cite{radford2021learning, zhou2022learning}. This is related to over-training on base classes and the catastrophic forgetting of domain-general knowledge~\cite{zhu2023prompt}. 

To address these challenges, we propose a mixture of soft prompt learning method. This method incorporates a routing module that selects the most suitable prompts for each instance. The selected prompts are then encoded by a text encoder to obtain several sets of class text features. These features are weighted and averaged to produce the final set of class text features, which are then compared with image features to calculate similarities. Conceptually, this process can be deemed as selecting the most compatible style prompts for each instance, thereby enhancing the system's adaptability and performance. 

For the router, we also propose a hard prompt guided gating loss to ensure it selects the soft prompts initialized from the hard prompt templates whose text features are the most similar to the image feature. This mechanism distills the knowledge of hard prompt templates into the router and encourages it to make more accurate and relevant selections.

Additionally, to mitigate the overfitting issue, we introduce semantically grouped text-level supervision. Each soft prompt corresponds to a set of manually designed templates (hard prompts), where the semantics within each set are relatively close. We use the token embeddings of one of the templates from each set as the initialization for each soft prompt. During training, the text features obtained by the text encoder for each soft prompt are constrained to stay close to the text features obtained from their corresponding  hard prompts. This ensures that the initial knowledge from the manual text templates is preserved and integrated into the soft prompts.  

We validated our method on 11 datasets, under the few-shot learning, domain generalization and base-to-new generalization from three main aspects. Our methods achieve improvements compared to existing baselines. We also designed ablation experiments to verify the contribution of different modules in our method to the performance improvement. 

In summary, our contributions are as follows:
\begin{itemize}[noitemsep, topsep=0pt]
    \item We propose a mixture of soft prompt learning method that incorporates a routing module to select the most suitable prompts for each instance.
    \item We introduce a hard prompt guided gating loss to ensure the router selects prompts based on their similarity to hard prompt templates, thus improving selection accuracy.
    \item We implement semantically grouped text-level supervision to maintain the initial knowledge from manual text templates and mitigate over-fitting.
    \item We validate our method on 11 datasets, demonstrating improvements in few-shot learning, domain generalization, and base-to-new generalization scenarios compared to existing baselines.
\end{itemize}

\section{Related Works}
\noindent{\bf Prompt Learning.} In the realm of vision-language models, prompt learning aims to bridge the gap between visual and textual representations more effectively. A pioneering work in this area is the CoOp (Context Optimization) model~\cite{zhou2022learning}, which optimizes the context of prompts to enhance the performance of models like CLIP~\cite{radford2021learning} in few-shot learning scenarios.

Researchers have also introduced the concept of a vision prompt~\cite{zang2022unified, khattak2023maple}, which involves appending learnable vectors to the inputs of a vision encoder, similar to text prompts. This approach can significantly enhance performance, although it also increases computational demands. In this paper, we focus exclusively on text-based prompts. In the future, our methodology could potentially be extended to include vision prompts.

Despite their success, most prompt learning methods trade-off between classification accuracy and robustness, e.g. in domain generalization or out-of-distribution (OOD) detection. A variety of methods have been developed to constrain the update of soft prompts using features from the original manual templates. These methods either directly restrict the gradient update direction or employ knowledge distillation. Among them, ProGrad~\cite{zhu2023prompt} prevents prompt tuning from forgetting general knowledge in VLMs by updating prompts only when their gradients align with the "general direction" represented by the KL loss gradient of a predefined prompt. LASP~\cite{bulat2022lasp} use grouped manual templates encoded feature as supervision to regularize the learning of the prompt. KgCoOp~\cite{yao2023visual} reduces the difference between the textual embeddings generated by learned prompts and those from hand-crafted prompts. We also incorporate this technique by distilling the knowledge from original text features into each expert soft prompt. Additionally, we apply gating regularization to distill prior knowledge from discrete text into the router. 

PLOT~\cite{chenplot} first explored to learn multiple comprehensive prompts to describe diverse characteristics of categories, using optimal transport to align visual and textual features. This method improves few-shot recognition tasks by applying a two-stage optimization strategy, demonstrating superior performance across various datasets compared to conventional prompt learning approaches. We in another way, use multiple prompts to capture the diverse styles in the dataset and learning to prompt in a sparse mixture of experts way.

\noindent{\bf Mixture of Experts.} The mixture of experts (MoE) framework~\cite{zhou2022mixture, masoudnia2014mixture}, initially introduced decades ago, has brought significant advancements for AI, especially with the advent of sparsely-gated MoE in transformer-based large language models~\cite{sukhbaatar2024branch,liu2024deepseek}. This framework allows different parts of a model, known as experts, to specialize in various tasks, engaging only relevant experts for a given input to maintain computational efficiency while leveraging specialized knowledge. A major issue of MoE is effectively balancing the load among different expert models, as poor load distribution can result in inefficiencies and unstable model performance~\cite{masoudnia2014mixture}.

\section{Method}
\subsection{Overview}
\begin{figure*}[t]
\centering
\includegraphics[width=1.\textwidth]{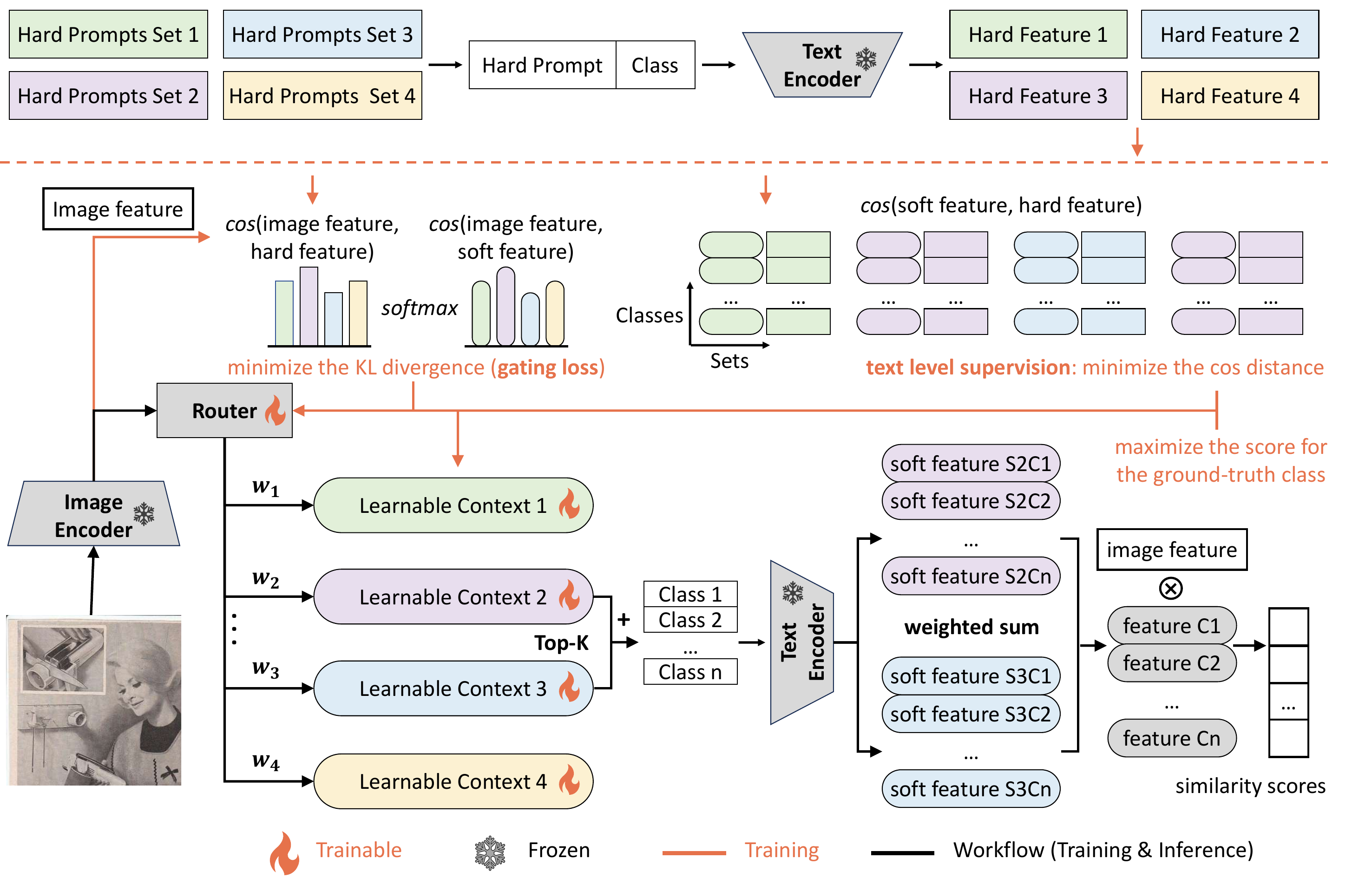} 
\caption{Overview of MoCoOp. The orange  lines signify the extra flow for training while the black lines are shared by training and inference. During inference, two soft prompts with the highest probabilities are selected and combined with the available classes for text encoding. The resulting text features are averaged and used for classification. During training, the hard prompt guided routing and semantically grouped text level supervision are introduced to supervise the router and soft prompts respectively. In our experiments, we set k to 2.}
\label{fig:framework}
\end{figure*}

As illustrated in Figure.~\ref{fig:framework}, during inference, an image is first processed by the CLIP image encoder to obtain an image feature. This feature is then routed to select the k soft prompts with the highest probabilities. These selected prompts are concatenated with the available classes and fed into the CLIP text encoder, resulting in k sets of class text features. These k sets are then averaged weighted by the router's gating distribution (after the softmax layer) to produce a single set of class text features. The final feature set is compared with the image feature to produce the classification logits. In this way, only k soft prompts are activated at a time, keeping the inference cost comparable to using a single prompt. 
During training, there are three parts of gradient flow.
First, we apply a cross entropy loss to the final classification probabilities with the ground truth label. Second, for the router, we calculate the similarity between the image feature and the text features from each hard prompt template set (using the average feature of all classes and all templates in the set). These similarities serve as a reference distribution. Then, a KL divergence objective function is used to align the router's gating distribution with this reference distribution. Finally, for the soft prompts, we use another cross entropy loss to ensure that each class's text feature from each soft prompt closely matches the corresponding class's feature from the associated hard prompt.
\subsection{Preliminary of CoOp}
Here we give a brief introduction of CoOp~\cite{zhou2022learning}, the pioneering work in prompt learning of VLMs.

\noindent{\bf Notation:}

First, here are some notations used in prompt learning of VLMs.
\begin{itemize}
    \item \( \mathbf{x} \): Input image
    \item \( \mathbf{p} \): Text prompt
    \item \( f_{\text{img}} \): CLIP image encoder
    \item \( f_{\text{txt}} \): CLIP text encoder
    \item \( \mathbf{h}_x = f_{\text{img}}(\mathbf{x}) \): Encoded image feature
    \item \( \mathbf{h}_p = f_{\text{txt}}(\mathbf{p}) \): Encoded text feature
    \item \( \mathbf{C} \): Context vectors (learnable parameters)
\end{itemize}

\noindent{\bf Prompt Representation.} The text prompt \( \mathbf{p} \) is represented as a sequence of tokens, including learnable context tokens and a class token. 
\[
\mathbf{p} = [\mathbf{C}, \text{CLASS}]
\]

The context tokens can also be placed after or around the class token.

\noindent{\bf Context:}
\begin{itemize}
    \item The context is learnable vectors \( \mathbf{C} = [\mathbf{c}_1, \mathbf{c}_2, \ldots, \mathbf{c}_M] \), where \( \mathbf{c}_i \in \mathbb{R}^d \) and \( M \) is the number of context tokens.
    \item All classes share the same context \( \mathbf{C} \) or each class \( c \) has its own context \( \mathbf{C}_c \).
\end{itemize}

\noindent{\bf Training Objective.} Given a dataset with images \( \{\mathbf{x}_i\} \) and corresponding labels \( \{y_i\} \), the goal is to find the optimal context vectors \( \mathbf{C} \) (or \( \mathbf{C}_c \) for class-specific context) by minimizing the cross-entropy loss:
\[
\mathcal{L} = - \sum_{i} \log \frac{\exp(\text{sim}(\mathbf{h}_x^i, \mathbf{h}_p^{y_i})/\tau)}{\sum_{c} \exp(\text{sim}(\mathbf{h}_x^i, \mathbf{h}_p^c)/\tau)}
\]
where
\begin{itemize}
    \item \( \mathbf{h}_x^i = f_{\text{img}}(\mathbf{x}_i) \) is the image feature for image \( i \).
    \item \( \mathbf{h}_p^c = f_{\text{txt}}([\mathbf{C}, \text{CLASS}_c]) \) is the text feature for class \( c \).
    \item \( \text{sim}(\cdot, \cdot) \) denotes a similarity function, such as cosine similarity.
    \item \(\tau\) is the temperature.
\end{itemize}

\noindent{\bf Optimization.} The context vectors \( \mathbf{C} \) are updated through backpropagation to minimize the loss \( \mathcal{L} \), while keeping the pre-trained parameters of \( f_{\text{img}} \) and \( f_{\text{txt}} \) fixed.

In summary, CoOp involves learning optimal context vectors \( \mathbf{C} \) for text prompts, which are used to synthesize classification weights for downstream tasks. This process automates prompt engineering and enhances the adaptability and performance of vision-language models like CLIP on various image recognition tasks.

\subsection{Mixture of Prompt Learning}
The essential idea of this work is to learn to prompt like mixture of experts. In LLMs, the router selects the top K experts for each input token.
Similarly, we use a router to select the top K contexts. Then 
the selected contexts are concatenated with the class names and encoded by the text encoder to obtain several sets of class features:
\begin{equation}
\mathbf{h}_{p_i} = f_{\text{txt}}([\mathbf{C}_i, \text{CLASS}])
\end{equation}
for \(i = 1, 2, \ldots, K\), where \( \mathbf{C}_i \) are the context vectors for the \(i\)-th selected prompt.

The features are then weighted and averaged to produce the final set of class features:
\begin{equation}
\mathbf{h}_p = \sum_{i=1}^{K} w_{\text{router}}^i \mathbf{h}_{p_i}
\end{equation}
where \(w_i\) are the weights assigned to each prompt feature.
A cross entropy loss is utilized to optimize these prompts:
\begin{equation}
\mathcal{L}_{\text{cls}} = - \sum_{i} \log \frac{\exp(\text{cos}(\mathbf{h}_x^i, \mathbf{h}_p^{y_i})/\tau)}{\sum_{c\in \mathcal{C}} \exp(\text{cos}(\mathbf{h}_x^i, \mathbf{h}_p^c)/\tau)}
\label{eq:loss_cls}
\end{equation}
\subsection{Hard Prompt Guided Routing}
Given \(G\) sets of hard prompts ($I_1, I_2, ... I_G$),  each concatenated with every class and encoded through the CLIP text encoder, we obtain \(G\) sets of hard text features for all classes. Specifically, for a hard prompt concatenated with a specific $\text{CLASS}_c$, the corresponding hard text features can be similarly obtained using the CLIP text encoder, resulting in:

\begin{equation}
\mathbf{h}_{c} = f_{\text{txt}}([\text{hard\_prompt}, \text{CLASS}_c])
\end{equation}

where \(c\) denotes the specific class.

These hard text features are then averaged to generate \(G\) group text features, each representing one of the \(G\) groups. Specifically, the group text feature \(\mathbf{h}_g\) for the \(g\)-th group is computed by averaging the hard text features for all classes and all templates within that group as:

\begin{equation}
\mathbf{h}_g = \frac{1}{|I_g|}\sum_{i \in I_g} \frac{1}{|\mathcal{C}|} \sum_{c \in \mathcal{C}} \mathbf{h}_{i,c}
\end{equation}

where \(\mathcal{C}\) represents the set of all classes, and \(\mathbf{h}_{i,c}\) represents the i-th hard text feature for class \(c\) in the \(g\)-th group.

The cosine similarity between the image feature \(\mathbf{v}\) and each group's text feature, is calculated.
The hard prompt guided gating distribution $W_\text{hard}$ is then derived by applying the softmax function to these similarity scores, expressed as:

\begin{align}
W_\text{hard} = \text{Softmax}\left(
\begin{array}{c}
\cos(\mathbf{h}_1, \mathbf{v}) \\
\cos(\mathbf{h}_2, \mathbf{v}) \\
\vdots \\
\cos(\mathbf{h}_G, \mathbf{v})
\end{array}
\right)
\end{align}

The router's output gating distribution is denoted by $W_\text{router}$. To ensure coherence between the two distributions, KL divergence is employed as a constraint, with the loss function defined as:

\begin{equation}
\mathcal{L}_{\text{router}} = D_{\text{KL}}(W_\text{router} \parallel W_\text{hard})
\end{equation}

\subsection{Semantically Grouped Text Level Supervision}
To mitigating the overfitting issue, we introduce semantically grouped text level supervision to allievating the overfitting issue. 

\begin{figure*}[thb!]
\centering
\includegraphics[width=1.\textwidth]{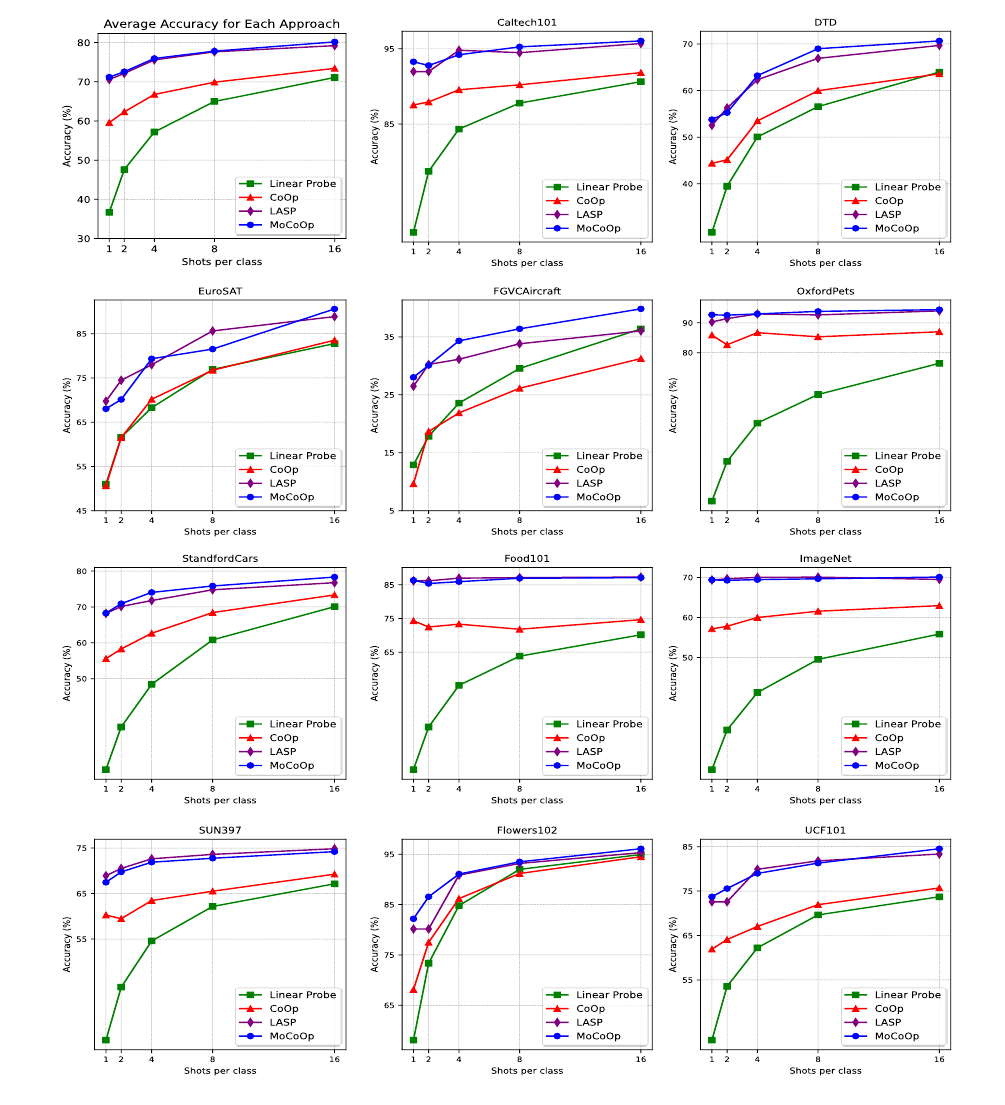} 
\caption{The few-shot learning results on 11 datasets. We plot the results across 1,2,4,8,16 shots. It can be seen that our MoCoOp consistently and significantly surpasses CoOp, LASP, and the Linear Probe approach across most datasets. This is evident in the average accuracy displayed in the top left corner. For LASP~\cite{bulat2022lasp}, we use our reproduced results.}
\label{fig:baseall}
\end{figure*}

The hard prompts are semantically grouped into G sets $I_1, I_2, ... I_G$. (See \ref{sec:template} for details). For each learnable soft prompt \(\mathbf{t}_g^s\) and its corresponding hard prompt group $I_g$, the probability of a class y filled in this prompt being classified as its proper class \(y\) is given by:

\begin{equation}
\begin{aligned}
&P(y|\mathbf{t}_g^s) = \frac{1}{|I_g|} \sum_{i \in I_g} \ P_i(y|\mathbf{t_g}^s)\\
&P_i(y|\mathbf{t_g}^s)=\frac{\exp \left( \cos \left( \mathbf{h}_{i, y}, f_\text{txt}([\mathbf{t}_g^s, y]) \right) / \tau \right)}
{\sum_{c \in \mathcal{C}} \exp \left( \cos \left( \mathbf{h}_{i, c}, f_\text{txt}([\mathbf{t}_g^s, c]) \right) / \tau \right)}
\end{aligned}
\label{eq:semantic-1}
\end{equation}

where $P_i(y|\mathbf{t_g}^s)$ is the possibility of $\mathbf{t_g}^s$ applied to class y be classified as the i-th hard template in $I_g$ applied to class y, \(\cos(\cdot, \cdot)\) denotes the cosine similarity, and \(\tau\) is a temperature parameter, $\mathcal{C}$ is the class set.

Next, we use the cross-entropy loss to minimize the distance between the encoded learnable soft prompts and the manually defined text prompts in the encoded space. The loss function can be expressed as:

\begin{equation}
\mathcal{L}_{\text{text}} =  -\frac{1}{G}\sum_{g=1}^{G}\sum_{c\in \mathcal{C}}\frac{1}{|\mathcal{C}|} \log P(c|\mathbf{t}^s_g)
\end{equation}


The overall training objective is
\begin{equation}
\mathcal{L} = \mathcal{L}_{\text{cls}} + \lambda_1 \mathcal{L}_{\text{router}} + \lambda_2 \mathcal{L}_{\text{text}}
\end{equation}
Where $\lambda_1$ and $\lambda_2$ are weights that balance the importance of each loss term.

\section{Experiment}

\begin{table*}[ht]
\centering
\footnotesize
\setlength{\tabcolsep}{0.9mm}
\resizebox{\textwidth}{!}
{
\begin{tabular}{lcccccccccccccccccc}
\toprule
\multirow{2}{*}{Dataset} & \multicolumn{3}{c}{CLIP} & \multicolumn{3}{c}{CoOp} & \multicolumn{3}{c}{CoCoOp} & \multicolumn{3}{c}{LASP} & \multicolumn{3}{c}{KgCoOp} & \multicolumn{3}{c}{MoCoOp (Ours)}\\
\cmidrule(lr){2-4} \cmidrule(lr){5-7} \cmidrule(lr){8-10} \cmidrule(lr){11-13} \cmidrule(lr){14-16}  \cmidrule(lr){17-19}
 & Base & New & H & Base & New & H & Base & New & H & Base & New & H & Base & New & H & Base & New & H\\
\midrule
Average & 70.25 & 74.22 & 71.57 & 82.64 & 68.00 & 74.02 & 80.47 & 71.69 & 75.42 & 83.18 & 76.11 & 79.48 & 73.63 & 76.90 & 83.22 & \textbf{83.32} & \textbf{77.34} & \textbf{80.17} \\
ImageNet & 72.43 & 68.14 & 70.22 & 76.46 & 66.31 & 71.00 & 75.98 & 70.43 & 73.11 & 76.25 & 71.17 & 73.62 & 75.83 & 69.96 & 71.89 & \textbf{76.52} & 69.2 & 72.67 \\
Caltech101 & 96.84 & 94.00 & 95.40 & 98.11 & 93.52 & 95.76 & 97.96 & 93.81 & 95.84 & 98.17 & 94.33 & 96.21 & 97.72 & 94.39 & 95.55 & \textbf{98.43} & \textbf{94.87} & \textbf{96.61} \\
OxfordPets & 91.17 & \textbf{97.26} & 94.11 & 94.24 & 96.66 & \textbf{95.44} & 95.20 & 97.69 & \textbf{96.43} & 95.73 & \textbf{97.87} & \textbf{96.79} & 94.65 & \textbf{97.76} & 96.18 & 95.59 & 96.64 & 96.11 \\
StanfordCars & 63.37 & 74.89 & 68.61 & 76.2 & 69.14 & 72.51 & 70.49 & 73.59 & 72.01 & 75.23 & 71.77 & 73.46 & 71.76 & 75.04 & 73.36 & \textbf{76.34} & 73.26 & \textbf{74.77} \\
Flowers102 & 72.08 & 77.80 & 74.82 & 97.63 & 69.55 & 81.35 & 94.87 & 71.75 & 81.64 & 97.17 & 73.53 & 83.71 & 95.00 & 74.73 & 83.65 & \textbf{97.18} & 77.21 & \textbf{86.05} \\
Food101 & 90.10 & 91.22 & 90.66 & 89.44 & 87.5 & 88.46 & 90.70 & \textbf{91.29} & 90.99 & 91.20 & \textbf{91.90} & \textbf{91.54} & 90.50 & \textbf{91.70} & 91.10 & 90.25 & 91.57 & 90.90 \\
FGVCAircraft & 27.19 & 36.29 & 31.07 & 39.24 & 30.49 & 34.23 & 33.41 & 23.71 & 27.73 & 38.05 & 33.20 & 35.46 & 36.21 & 33.55 & 34.83 & \textbf{38.78} & \textbf{38.09} & \textbf{38.43} \\
SUN397 & 69.36 & 75.35 & 72.22 & 80.85 & 68.34 & 74.06 & 79.74 & 76.86 & 78.27 & 80.70 & 79.30 & \textbf{80.00} & 80.29 & 76.53 & 78.36 & \textbf{81.43} & 77.45 & 79.39 \\
DTD & 53.24 & 59.90 & 56.36 & 80.17 & 47.54 & 59.67 & 77.01 & 56.00 & 64.85 & 81.10 & 62.57 & 70.64 & 77.55 & 54.99 & 64.35 & \textbf{81.94} & 60.99 & 69.93 \\
EuroSAT & 56.48 & 64.05 & 60.03 & 91.54 & 54.44 & 68.15 & 87.49 & 60.04 & 71.11 & \textbf{95.00} & 83.37 & 88.86 & \textbf{95.64} & 64.34 & 76.93 & 94.79 & 85.18 & \textbf{89.73} \\
UCF101 & 70.53 & 77.50 & 73.82 & 85.14 & 64.47 & 73.57 & 82.33 & 73.45 & 77.63 & 85.53 & 78.20 & 81.70 & 82.89 & 76.67 & 79.66 & \textbf{85.28} & 79.31 & \textbf{82.17} \\
\bottomrule
\end{tabular}
}
\caption{The comparison with baselines on novel class prediction. H is the harmonic mean of the test accuracy on base and new class. The best results are marked in bold font.}
\label{tab:base2new}
\end{table*}

\begin{table*}[h!]
\centering
  \resizebox{\textwidth}{!}
  {
\begin{tabular}{lcccccccccccc}
\toprule
& \multicolumn{3}{c}{\textbf{Caltech101}} & \multicolumn{3}{c}{\textbf{EuroSAT}} & \multicolumn{3}{c}{\textbf{UCF101}}&  \multicolumn{3}{c}{\textbf{Flowers102}}\\ 
\cmidrule(lr){2-4} \cmidrule(lr){5-7} \cmidrule(lr){8-10} \cmidrule{11-13}
& \textbf{Base} & \textbf{New} & \textbf{H} & \textbf{Base} & \textbf{New} & \textbf{H} & \textbf{Base} & \textbf{New} & \textbf{H} & \textbf{Base} & \textbf{New} & \textbf{H} \\ 
\midrule
Baseline& 95.40 & 98.11 & 93.52& 91.54 & 54.44 & 68.15& 85.14 & 64.47 & 73.57& 97.63 & 69.55 & 81.35 \\
+ MoE &98.38&92.03&95.10&94.90&58.79&72.60&85.78& 69.50&76.79&97.63&70.64&81.97 \\ 
+$L_\text{router}$& 98.39&92.47&95.34&95.17&57.05&71.34&86.97&73.88&79.89&97.34&72.77&83.28\\ 
+$L_\text{text}$ &98.43& 94.87& 96.61& 94.79& 85.18& 89.73&85.28& 79.31& 82.17&97.18& 77.21& 86.05 \\ 
\bottomrule
\end{tabular}
}
\caption{Component analysis. We sequentially add the components MoE, $L_\text{router}$ and $L_\text{text}$. Our baseline is CoOp~\cite{zhou2022learning}}
\label{tab:component}
\end{table*}


\noindent{\bf Settings:} We conduct experiments under three settings: base to new generalization, few-shot learning, and domain generalization. For base to new generalization, we train on the base class and test on both the base class and new class. For few-shot learning, we train and test on all classes.And domain generalization refers to training on ImageNet and testing on other datasets. The few-shot capability reflects the method's fitting ability, while base-to-new generalization and domain generalization can measure the model's robustness.

\noindent{\bf Implementation Details:} We build our framework based on LASP~\cite{bulat2022lasp}. For each expert, we use different context positions depending on the handcrafted template object used to initialize it. We used 4 to 20 experts. The number of experts and corresponding templates varies for datasets. For example, for FGVC\_Aircraft, we use the template "a photo of a \{\}, a type of aircraft." For the Oxford\_Flowers dataset, we use "a photo of a \{\}, a type of flower." Generally, a unique template for the dataset is combined with some general templates like "a photo of a {}". Since ImageNet covers a wide range of categories, we use 20 groups of templates. Specific templates can be found in the appendix~\ref{sec:template}. Based on existing studies, we use ViT-B/16 as the backbone. Specifically, we use the publicly available CLIP-ViT-B/16 models (https://github.com/openai/CLIP). The resolution of CLIP's feature map is 14 × 14 for CLIP-ViT-B/16. The $\lambda_1$ and $\lambda_2$ is set as 1. and 5. respectively. The $\tau$ in Eq.\ref{eq:loss_cls}  and Eq.\ref{eq:semantic-1} is set to 0.07.
For base-to-new generalization, we use virtual classes during training following LASP~\cite{bulat2022lasp} by incorporating new classes as text-level supervision. This approach helps mitigate overfitting to some extent.

\noindent{\bf Evaluation Metrics:} For few-shot experiments, we use top-1 accuracy. For base to new generalization, we evaluate by base class accuracy, new class accuracy, and the harmonic mean of base and new classes.

\noindent{\bf Training:} Our training schedule is consistent with LASP~\cite{bulat2022lasp}, and both training and testing are conducted on four NVIDIA GeForce RTX 3090 GPUs.

\noindent{\bf Baselines:} In the few-shot experiment, we compared with Linear Probe, CoOp~\cite{zhou2022learning}, PLOT~\cite{chenplot}, and LASP~\cite{bulat2022lasp}.
In the base-to-new generalization experiment, we compare with CoOp~\cite{zhou2022learning}, CoCoOp~\cite{zhou2022conditional}, KgCoOp~\cite{yao2023visual} and LASP~\cite{bulat2022lasp}. 
Note that CoOp~\cite{zhou2022learning}, KgCoOp~\cite{yao2023visual}, LASP~\cite{bulat2022lasp}, PLOT~\cite{chenplot} are textual only methods while CoCoOp~\cite{zhou2022conditional} is instance-conditioned. 
Textual-only methods typically have poorer generalization to unseen classes within the same task, even lagging behind the original CLIP on some datasets. Instance-conditioned methods improves the generalization by generating different contexts based on various image visual features, and then obtain different text features through the CLIP text encoder. Therefore, they require significant computational resources. Our method, MoCoOp, also partially relies on visual information but does not generate new contexts. Instead, it combines different text features for different images, thus eliminating the heavy computational cost of the text encoder during inference.

\noindent{\bf Dataset:}
Following previous studies~\cite{zhou2022learning, zhou2022conditional, chenplot, yao2023visual,bulat2022lasp}, we primarily evaluate the accuracy of our approach  across a total of 11 datasets. The datasets used include: ImageNet~\cite{deng2009imagenet}, Caltech101~\cite{fei2004learning}, Oxford-Pets~\cite{parkhi2012cats}, Stanford Cars~\cite{krause20133d}, Flowers102~\cite{nilsback2008automated}, Food101~\cite{bossard2014food}, FGVC Aircraft~\cite{maji2013fine}, SUN397~\cite{xiao2010sun}, DTD~\cite{cimpoi2014describing}, EuroSAT~\cite{helber2019eurosat}, and UCF-101~\cite{soomro2012ucf101}.

\subsection{Main Results}
Here we show the results of few-shot experiments and base-to-new generalization. The domain generalization results can be found in Appendix.~\ref{sec:dg}
\subsubsection{Results of Few-shot experiment}
In the Figure~\ref{fig:baseall}, we plot the performance curves of our MoCoOp and the baselines across 11 datasets for various shots, along with the average accuracies of all datasets. It can be seen that our method achieves the best results in most cases. The performance on ImageNet is average, possibly because other methods utilized all 39 hand-crafted templates, whereas we need to control the number of groups and selected only a portion. Since ImageNet contains images with diverse styles, using only a subset of templates might not have been sufficient.

\subsubsection{Results of Base-to-New Generalization}
In the Table~\ref{tab:base2new}, we list the comparison results of MoCoOp and several baselines. It can be seen that our method surpasses the baselines in generalization ability on most datasets, especially compared to LASP~\cite{bulat2022lasp}. The improvement can be attributed to the utilization of multiple prompts and the semantically grouped text supervision.

\subsection{Ablations}
\subsubsection{Component Analysis. }
Table.\ref{tab:component} presents the performance as we progressively include components. Our baseline is CoOp ~\cite{zhou2022learning}. As can be seen in Table.~\ref{tab:component}, adding MoE alone has already achieved significant improvement. Adding hard prompt guided routing provides a slight improvement, while incorporating semantically grouped text supervision brings a huge enhancement.

\section{Conclusion}
In this work, we introduce a novel mixture of prompt learning method for vision-language models, addressing key challenges such as dataset style variations and overfitting. Our approach employs a routing module to dynamically select the most suitable prompts for each instance, enhancing adaptability and performance. We also propose a hard prompt guided gating loss and semantically grouped text-level supervision, which help maintain initial knowledge and mitigate overfitting. Our method demonstrate significant improvements across multiple datasets in few-shot learning, domain generalization, and base-to-new generalization scenarios. Future work could explore extending this methodology to include vision prompts or instance-conditioned contexts for further enhancements. Another direction could be using ChatGPT for generating and grouping hard prompt templates.
\section{Limitations}
While our MoCoOp demonstrates improvements across various tasks, there are two limitations.
First, despite the sparse gating of soft prompts, the training cost and memory usage remain high compared to single prompt methods. This can be a constraint in resource-limited environments, especially when dealing with large-scale datasets.
Second, templates require manual grouping based on their semantics, potentially introducing human bias that could affect the model's performance. To enhance efficiency and accuracy, developing automated grouping algorithms may be necessary in the future.

\bibliography{ref}

\begin{thebibliography}{30}
\providecommand{\natexlab}[1]{#1}

\bibitem[{Bossard et~al.(2014)Bossard, Guillaumin, and Van~Gool}]{bossard2014food}
Lukas Bossard, Matthieu Guillaumin, and Luc Van~Gool. 2014.
\newblock Food-101--mining discriminative components with random forests.
\newblock In \emph{Computer vision--ECCV 2014: 13th European conference, zurich, Switzerland, September 6-12, 2014, proceedings, part VI 13}, pages 446--461. Springer.

\bibitem[{Bulat and Tzimiropoulos(2022)}]{bulat2022lasp}
Adrian Bulat and Georgios Tzimiropoulos. 2022.
\newblock Lasp: Text-to-text optimization for language-aware soft prompting of vision \& language models.
\newblock \emph{arXiv preprint arXiv:2210.01115}.

\bibitem[{Chen et~al.()Chen, Yao, Song, Li, Rao, and Zhang}]{chenplot}
Guangyi Chen, Weiran Yao, Xiangchen Song, Xinyue Li, Yongming Rao, and Kun Zhang.
\newblock Plot: Prompt learning with optimal transport for vision-language models.
\newblock In \emph{The Eleventh International Conference on Learning Representations}.

\bibitem[{Cimpoi et~al.(2014)Cimpoi, Maji, Kokkinos, Mohamed, and Vedaldi}]{cimpoi2014describing}
Mircea Cimpoi, Subhransu Maji, Iasonas Kokkinos, Sammy Mohamed, and Andrea Vedaldi. 2014.
\newblock Describing textures in the wild.
\newblock In \emph{Proceedings of the IEEE conference on computer vision and pattern recognition}, pages 3606--3613.

\bibitem[{Crowson et~al.(2022)Crowson, Biderman, Kornis, Stander, Hallahan, Castricato, and Raff}]{crowson2022vqgan}
Katherine Crowson, Stella Biderman, Daniel Kornis, Dashiell Stander, Eric Hallahan, Louis Castricato, and Edward Raff. 2022.
\newblock Vqgan-clip: Open domain image generation and editing with natural language guidance.
\newblock In \emph{European Conference on Computer Vision}, pages 88--105. Springer.

\bibitem[{Deng et~al.(2009)Deng, Dong, Socher, Li, Li, and Fei-Fei}]{deng2009imagenet}
Jia Deng, Wei Dong, Richard Socher, Li-Jia Li, Kai Li, and Li~Fei-Fei. 2009.
\newblock Imagenet: A large-scale hierarchical image database.
\newblock In \emph{2009 IEEE conference on computer vision and pattern recognition}, pages 248--255. Ieee.

\bibitem[{Eslami et~al.(2021)Eslami, de~Melo, and Meinel}]{eslami2021does}
Sedigheh Eslami, Gerard de~Melo, and Christoph Meinel. 2021.
\newblock Does clip benefit visual question answering in the medical domain as much as it does in the general domain?
\newblock \emph{arXiv preprint arXiv:2112.13906}.

\bibitem[{Fei-Fei et~al.(2004)Fei-Fei, Fergus, and Perona}]{fei2004learning}
Li~Fei-Fei, Rob Fergus, and Pietro Perona. 2004.
\newblock Learning generative visual models from few training examples: An incremental bayesian approach tested on 101 object categories.
\newblock In \emph{2004 conference on computer vision and pattern recognition workshop}, pages 178--178. IEEE.

\bibitem[{Helber et~al.(2019)Helber, Bischke, Dengel, and Borth}]{helber2019eurosat}
Patrick Helber, Benjamin Bischke, Andreas Dengel, and Damian Borth. 2019.
\newblock Eurosat: A novel dataset and deep learning benchmark for land use and land cover classification.
\newblock \emph{IEEE Journal of Selected Topics in Applied Earth Observations and Remote Sensing}, 12(7):2217--2226.

\bibitem[{Hendrycks et~al.(2021{\natexlab{a}})Hendrycks, Basart, Mu, Kadavath, Wang, Dorundo, Desai, Zhu, Parajuli, Guo et~al.}]{hendrycks2021many}
Dan Hendrycks, Steven Basart, Norman Mu, Saurav Kadavath, Frank Wang, Evan Dorundo, Rahul Desai, Tyler Zhu, Samyak Parajuli, Mike Guo, et~al. 2021{\natexlab{a}}.
\newblock The many faces of robustness: A critical analysis of out-of-distribution generalization.
\newblock In \emph{Proceedings of the IEEE/CVF international conference on computer vision}, pages 8340--8349.

\bibitem[{Hendrycks et~al.(2021{\natexlab{b}})Hendrycks, Zhao, Basart, Steinhardt, and Song}]{hendrycks2021natural}
Dan Hendrycks, Kevin Zhao, Steven Basart, Jacob Steinhardt, and Dawn Song. 2021{\natexlab{b}}.
\newblock Natural adversarial examples.
\newblock In \emph{Proceedings of the IEEE/CVF conference on computer vision and pattern recognition}, pages 15262--15271.

\bibitem[{Khattak et~al.(2023)Khattak, Rasheed, Maaz, Khan, and Khan}]{khattak2023maple}
Muhammad~Uzair Khattak, Hanoona Rasheed, Muhammad Maaz, Salman Khan, and Fahad~Shahbaz Khan. 2023.
\newblock Maple: Multi-modal prompt learning.
\newblock In \emph{Proceedings of the IEEE/CVF Conference on Computer Vision and Pattern Recognition}, pages 19113--19122.

\bibitem[{Krause et~al.(2013)Krause, Stark, Deng, and Fei-Fei}]{krause20133d}
Jonathan Krause, Michael Stark, Jia Deng, and Li~Fei-Fei. 2013.
\newblock 3d object representations for fine-grained categorization.
\newblock In \emph{Proceedings of the IEEE international conference on computer vision workshops}, pages 554--561.

\bibitem[{Liu et~al.(2024)Liu, Feng, Wang, Wang, Liu, Zhao, Dengr, Ruan, Dai, Guo et~al.}]{liu2024deepseek}
Aixin Liu, Bei Feng, Bin Wang, Bingxuan Wang, Bo~Liu, Chenggang Zhao, Chengqi Dengr, Chong Ruan, Damai Dai, Daya Guo, et~al. 2024.
\newblock Deepseek-v2: A strong, economical, and efficient mixture-of-experts language model.
\newblock \emph{arXiv preprint arXiv:2405.04434}.

\bibitem[{Maji et~al.(2013)Maji, Rahtu, Kannala, Blaschko, and Vedaldi}]{maji2013fine}
Subhransu Maji, Esa Rahtu, Juho Kannala, Matthew Blaschko, and Andrea Vedaldi. 2013.
\newblock Fine-grained visual classification of aircraft.
\newblock \emph{arXiv preprint arXiv:1306.5151}.

\bibitem[{Masoudnia and Ebrahimpour(2014)}]{masoudnia2014mixture}
Saeed Masoudnia and Reza Ebrahimpour. 2014.
\newblock Mixture of experts: a literature survey.
\newblock \emph{Artificial Intelligence Review}, 42:275--293.

\bibitem[{Nilsback and Zisserman(2008)}]{nilsback2008automated}
Maria-Elena Nilsback and Andrew Zisserman. 2008.
\newblock Automated flower classification over a large number of classes.
\newblock In \emph{2008 Sixth Indian conference on computer vision, graphics \& image processing}, pages 722--729. IEEE.

\bibitem[{Parkhi et~al.(2012)Parkhi, Vedaldi, Zisserman, and Jawahar}]{parkhi2012cats}
Omkar~M Parkhi, Andrea Vedaldi, Andrew Zisserman, and CV~Jawahar. 2012.
\newblock Cats and dogs.
\newblock In \emph{2012 IEEE conference on computer vision and pattern recognition}, pages 3498--3505. IEEE.

\bibitem[{Radford et~al.(2021)Radford, Kim, Hallacy, Ramesh, Goh, Agarwal, Sastry, Askell, Mishkin, Clark et~al.}]{radford2021learning}
Alec Radford, Jong~Wook Kim, Chris Hallacy, Aditya Ramesh, Gabriel Goh, Sandhini Agarwal, Girish Sastry, Amanda Askell, Pamela Mishkin, Jack Clark, et~al. 2021.
\newblock Learning transferable visual models from natural language supervision.
\newblock In \emph{International conference on machine learning}, pages 8748--8763. PMLR.

\bibitem[{Recht et~al.(2019)Recht, Roelofs, Schmidt, and Shankar}]{recht2019imagenet}
Benjamin Recht, Rebecca Roelofs, Ludwig Schmidt, and Vaishaal Shankar. 2019.
\newblock Do imagenet classifiers generalize to imagenet?
\newblock In \emph{International conference on machine learning}, pages 5389--5400. PMLR.

\bibitem[{Soomro et~al.(2012)Soomro, Zamir, and Shah}]{soomro2012ucf101}
Khurram Soomro, Amir~Roshan Zamir, and Mubarak Shah. 2012.
\newblock Ucf101: A dataset of 101 human actions classes from videos in the wild.
\newblock \emph{arXiv preprint arXiv:1212.0402}.

\bibitem[{Sukhbaatar et~al.(2024)Sukhbaatar, Golovneva, Sharma, Xu, Lin, Rozi{\`e}re, Kahn, Li, Yih, Weston et~al.}]{sukhbaatar2024branch}
Sainbayar Sukhbaatar, Olga Golovneva, Vasu Sharma, Hu~Xu, Xi~Victoria Lin, Baptiste Rozi{\`e}re, Jacob Kahn, Daniel Li, Wen-tau Yih, Jason Weston, et~al. 2024.
\newblock Branch-train-mix: Mixing expert llms into a mixture-of-experts llm.
\newblock \emph{arXiv preprint arXiv:2403.07816}.

\bibitem[{Wang et~al.(2019)Wang, Ge, Lipton, and Xing}]{wang2019learning}
Haohan Wang, Songwei Ge, Zachary Lipton, and Eric~P Xing. 2019.
\newblock Learning robust global representations by penalizing local predictive power.
\newblock \emph{Advances in Neural Information Processing Systems}, 32.

\bibitem[{Xiao et~al.(2010)Xiao, Hays, Ehinger, Oliva, and Torralba}]{xiao2010sun}
Jianxiong Xiao, James Hays, Krista~A Ehinger, Aude Oliva, and Antonio Torralba. 2010.
\newblock Sun database: Large-scale scene recognition from abbey to zoo.
\newblock In \emph{2010 IEEE computer society conference on computer vision and pattern recognition}, pages 3485--3492. IEEE.

\bibitem[{Yao et~al.(2023)Yao, Zhang, and Xu}]{yao2023visual}
Hantao Yao, Rui Zhang, and Changsheng Xu. 2023.
\newblock Visual-language prompt tuning with knowledge-guided context optimization.
\newblock In \emph{Proceedings of the IEEE/CVF conference on computer vision and pattern recognition}, pages 6757--6767.

\bibitem[{Zang et~al.(2022)Zang, Li, Zhou, Huang, and Loy}]{zang2022unified}
Yuhang Zang, Wei Li, Kaiyang Zhou, Chen Huang, and Chen~Change Loy. 2022.
\newblock Unified vision and language prompt learning.
\newblock \emph{arXiv preprint arXiv:2210.07225}.

\bibitem[{Zhou et~al.(2022{\natexlab{a}})Zhou, Yang, Loy, and Liu}]{zhou2022conditional}
Kaiyang Zhou, Jingkang Yang, Chen~Change Loy, and Ziwei Liu. 2022{\natexlab{a}}.
\newblock Conditional prompt learning for vision-language models.
\newblock In \emph{Proceedings of the IEEE/CVF conference on computer vision and pattern recognition}, pages 16816--16825.

\bibitem[{Zhou et~al.(2022{\natexlab{b}})Zhou, Yang, Loy, and Liu}]{zhou2022learning}
Kaiyang Zhou, Jingkang Yang, Chen~Change Loy, and Ziwei Liu. 2022{\natexlab{b}}.
\newblock Learning to prompt for vision-language models.
\newblock \emph{International Journal of Computer Vision}, 130(9):2337--2348.

\bibitem[{Zhou et~al.(2022{\natexlab{c}})Zhou, Lei, Liu, Du, Huang, Zhao, Dai, Le, Laudon et~al.}]{zhou2022mixture}
Yanqi Zhou, Tao Lei, Hanxiao Liu, Nan Du, Yanping Huang, Vincent Zhao, Andrew~M Dai, Quoc~V Le, James Laudon, et~al. 2022{\natexlab{c}}.
\newblock Mixture-of-experts with expert choice routing.
\newblock \emph{Advances in Neural Information Processing Systems}, 35:7103--7114.

\bibitem[{Zhu et~al.(2023)Zhu, Niu, Han, Wu, and Zhang}]{zhu2023prompt}
Beier Zhu, Yulei Niu, Yucheng Han, Yue Wu, and Hanwang Zhang. 2023.
\newblock Prompt-aligned gradient for prompt tuning.
\newblock In \emph{Proceedings of the IEEE/CVF International Conference on Computer Vision}, pages 15659--15669.

\end{thebibliography}
\appendix
\label{sec:appendix}
\section{Groups of Hard Prompt Templates}
\label{sec:template}

Here is the groups of hard prompt templates.
\begin{lstlisting}
   [
    # Photos of flowers
     "a photo of a {}, a type of flower.",
    # Photos of people doing activities
        "a photo of a person doing {}.",
    # Satellite photos
        "a centered satellite photo of {}.",
    # Photos of aircraft
        "a photo of a {}, a type of aircraft.",
    # "Itap" (I took a picture) photos
        "itap of a {}.",
        "itap of the {}.",
    # Photos of large objects
        "a photo of the large {}.",
        "a photo of a large {}.",
    # Art and renditions
        "art of the {}.",
        "a rendering of a {}.",
        "a rendering of the {}.",
        "a rendition of the {}.",
    # Photos of small objects
        "a photo of the small {}.",
    # General photo prompts
        "a photo of a {}.",
        "a photo of the {}.",
        "a photo of many {}.",
    # Low resolution and pixelated photos
        "a low resolution photo of the {}.",
        "a low resolution photo of a {}.",
        "a pixelated photo of the {}.",
        "a pixelated photo of a {}.",
        "a jpeg corrupted photo of the {}.",
        "a blurry photo of a {}.",
        "a bad photo of the {}.",
    # Cropped photos
        "a cropped photo of the {}.", 
        "a cropped photo of a {}.", 
    # Bright photos
        "a bright photo of the {}.", 
    # Good quality photos
        "a good photo of the {}.",
        "a good photo of a {}.",
    # Close-up photos
        "a close-up photo of the {}.",
    # Jpeg corrupted photos
    # Blurry photos
    # Clean objects
        "a photo of the clean {}.",
    # Video game screenshots
        "a {} in a video game.",
    # Hard to see objects
        "a photo of the hard to see {}.",
    # Bad quality photos
    # Origami photos
        "a origami {}.",
    # Texture photos
        "{} texture.",
    ]
\end{lstlisting}
\section{Results of Domain Generalization}
\label{sec:dg}
\begin{table}[th]
\footnotesize
\setlength{\tabcolsep}{0.9mm}
\centering
{
\begin{tabular}{lccccc}
\toprule
\textbf{Method} & \multicolumn{1}{c}{\textbf{Source}} & \multicolumn{4}{c}{\textbf{Target}} \\
\cmidrule(r){2-2} \cmidrule(r){3-6}
& ImageNet & -R & -A & -Sketch & -V2 \\
\midrule
CLIP & 66.73 & 73.96 & 47.77 & 46.15 & 60.83 \\
CoOp & 71.51 & 75.21 & 49.71 & 47.99 & 64.20 \\
CoCoOp & 71.02 & 76.18 & 50.63 & 48.75 & 64.07 \\
ProGrad & 72.24 & 74.58 & 49.39 & 47.63 & 64.73 \\
KgCoOp & 71.20 & 76.70 & 50.69 & 48.97 & 64.10 \\
LASP & 69.49 & 75.54 & 47.08 & 47.59 & 62.52 \\
MoCoOp (Ours) &70.08 &75.88 &48.97 &46.50 &61.31 \\
\bottomrule
\end{tabular}
}
\caption{Comparisons on robustness to domain shift. All methods are trained on 16 shots per class of ImageNet and tested on ImageNet-R, ImageNet-A, ImageNet-Sketch and ImageNet-V2.  For LASP~\cite{bulat2022lasp}, we use our reproduced results.}
\label{tab:dg}
\end{table}
\begin{table}[thb]
\centering
\footnotesize
\setlength{\tabcolsep}{0.9mm}
\begin{tabular}{lcccccccccccc}
\toprule
& \multicolumn{3}{c}{\textbf{K=2}} & \multicolumn{3}{c}{\textbf{K=3}} & \multicolumn{3}{c}{\textbf{K=4}} \\ 
\cmidrule(lr){2-4} \cmidrule(lr){5-7} \cmidrule(lr){8-10} \cmidrule{11-13}
& \textbf{Base} & \textbf{New} & \textbf{H} & \textbf{Base} & \textbf{New} & \textbf{H} & \textbf{Base} & \textbf{New} & \textbf{H} \\ 
\midrule
Caltech101& 98.43& 94.87& 96.61 &98.39&94.43&96.37&98.00&94.87&96.41\\ 
EuroSAT& 94.48& 77.02& 84.75&94.38&75.0&83.58&94.02&74.36&83.04\\ 
UCF101&85.28& 79.31& 82.17&84.23&68.63&75.63&86.40&79.23&82.66 \\ 
Flowers102&97.18& 77.21& 86.05&98.10&71.42&82.66& 97.34&76.67&85.78\\
\bottomrule
\end{tabular}
\caption{Ablations of the number of selected experts.}
\label{tab:topk}
\end{table}

We evaluate our MoCoOp in the domain generalization setting. This evaluate the robustness to domain shift. We train on 16 shots of ImageNet and test on ImageNet-R~\cite{hendrycks2021many}, ImageNet-A~\cite{hendrycks2021natural}, ImageNet-Sketch~\cite{wang2019learning}, ImageNet-V2~\cite{recht2019imagenet}. As seen in Table~\ref{tab:dg}, the results are comparable with LASP~\cite{bulat2022lasp}.

\section{Additional Ablations}
\subsubsection{The number of experts selected by the router}
We also show the effect of the number of experts selected on the performance. As seen in Table~\ref{tab:topk}, using top 2 prompts is the best in most cases.

\end{document}